\documentclass[conference]{IEEEtran}
\usepackage{times}

\usepackage[numbers]{natbib}
\usepackage{multicol}
\usepackage[bookmarks=true]{hyperref}
\usepackage{graphicx}
\usepackage{wasysym}

\usepackage{color}
\definecolor{CommentRed}{rgb}{0.7,0,0}
\definecolor{CommentBlue}{rgb}{0,0,0.7}
\definecolor{CommentGreen}{rgb}{0,0.7,0}
\definecolor{CommentOrange}{rgb}{1.0,0.5,0.0}

\newcommand{\peter}[1] {{\color{CommentBlue} { PETER: \textbf{#1}}}}

\newcommand{\state}{y}
\newcommand{\hidden}{h}

\newcommand{\observation}{x}

\newcommand{\graveyard}[1]{}

\begin{document}

% paper title
\title{End-to-End Tracking and Semantic Segmentation\\Using Recurrent Neural Networks}

\author{\authorblockN{Peter Ondr\'{u}\v{s}ka}
\authorblockA{Mobile Robotics Group\\
University of Oxford, UK\\
ondruska@robots.ox.ac.uk}
\and
\authorblockN{Julie Dequaire}
\authorblockA{Mobile Robotics Group\\
University of Oxford, UK\\
julie@robots.ox.ac.uk}
\and
\authorblockN{Dominic Zeng Wang}
\authorblockA{Mobile Robotics Group\\
University of Oxford, UK\\
dominic@robots.ox.ac.uk}
\and
\authorblockN{Ingmar Posner}
\authorblockA{Mobile Robotics Group\\
University of Oxford, UK\\
ingmar@robots.ox.ac.uk}
}

\maketitle

\begin{abstract}
In this work we present a novel end-to-end framework for tracking and classifying a robot's surroundings in complex, dynamic and only partially observable real-world environments. The approach deploys a recurrent neural network to filter an input stream of raw laser measurements in order to directly infer object locations, along with their identity in both visible and occluded areas. To achieve this we first train the network using unsupervised \textit{Deep Tracking}, a recently proposed theoretical framework for end-to-end space occupancy prediction. We show that by learning to track on a large amount of unsupervised data, the network creates a rich internal representation of its environment which we in turn exploit through the principle of inductive transfer of knowledge to perform the task of it's semantic classification. As a result, we show that only a small amount of labelled data suffices to steer the network towards mastering this additional task. Furthermore we propose a novel recurrent neural network architecture specifically tailored to tracking and semantic classification in real-world robotics applications. We demonstrate the tracking and classification performance of the method on real-world data collected at a busy road junction. Our evaluation shows that the proposed end-to-end framework compares favourably to a state-of-the-art, model-free tracking solution and that it outperforms a conventional one-shot training scheme for semantic classification.
\end{abstract}

\IEEEpeerreviewmaketitle

\graveyard{
\peter{Key things to explain in the paper:
\begin{itemize}
\item Implicit tracking vs. Explicit tracking (Introduction) - predicting occupancy or another consequences of presence of objects instead of objects themselves hoping it will learn to track as a subtask. This is similar to semantic segmentation vs. object detection.
\item End-to-end learning vs. Multi-stage pipeline (Introduction).
\item Dynamic occupancy grid vs. Static occupancy grid (Problem Formulation).
\item Occupancy and Semantic Information. (Problem Formulation)
\item DeepTracking and time-predictive training. (this is the principle behind the magic).
\item New network schema and multi-task training.
\item New neural network architecture.
\item Predicting future.
\end{itemize}
}
}

\section{Introduction}

\graveyard{
\begin{itemize}
\item Application: Model-free tracking in occluded scenes.
\item Method: End-to-End Recurrent Neural Networks and Deep Learning.
\item Implementation: Neural network design.
\item Extension: Inductive Transfer to easily solve another related task.
\end{itemize}

Introduction talk about:
\begin{itemize}
\item robot perception and their challenges.
\item pitch: what is it, how is it useful, who will use it, how is it novel, why we do it now and not 10 years ago.
\end{itemize}
}

Complete and accurate situational awareness in complex, dynamic environments is a pivotal requirement for successful and safe robot operation. Often, however, this remains an elusive goal due to the limited field of view of the robot's on-board sensors and to the complex and usually wide-ranging occlusions encountered. This limitation can impose significant challenges on the planner and may lead to otherwise unnecessarily conservative robot behaviour~\cite{Richter2015}.

Object detection and tracking modules specifically addressing this problem are ubiquitous in robotics. Commonly, however, they feature multiple individual data-processing steps designed and optimised separately from one another.
Traditional model-free approaches~\cite{Vu2007,Yang2011,Wang01062015} make few assumptions with regards to the objects involved, such as their shapes or semantic characteristics, but they are often not robust. Model-based approaches~\cite{Arras2007,Petrovskaya2009,Zhao1998} on the other hand limit the generality of these frameworks and often require separate object segmentation and classification steps. 

\begin{figure}[t]
\centering
\includegraphics[width=80mm]{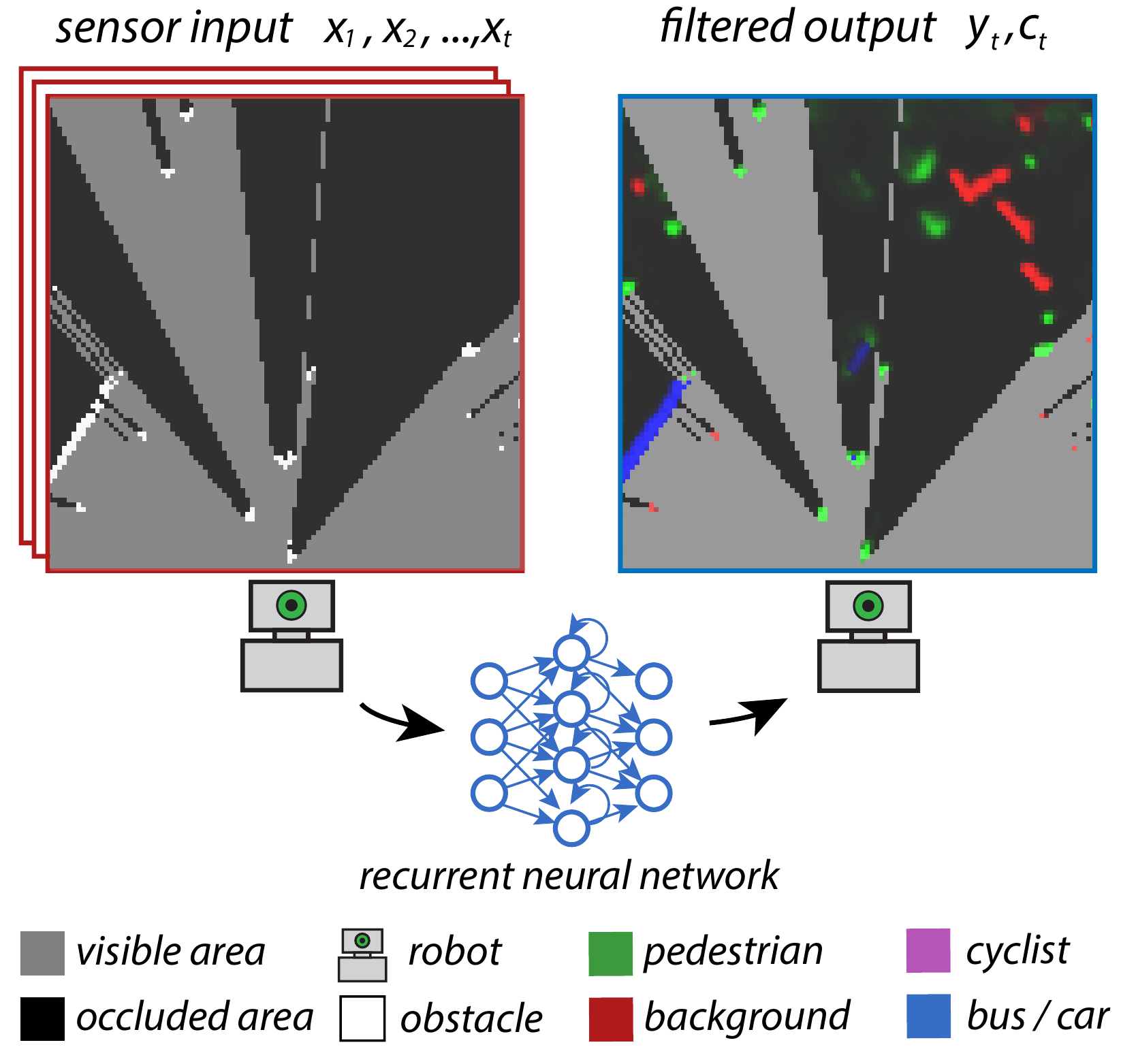}
\caption{Typical output of the proposed system capturing the situation around the robot in the form of a semantic map. The stream of raw sensor data is filtered by a recurrent neural network and produces classification of both directly visible and occluded space into one of several semantic classes.}
\label{fig:intro}
\end{figure}

In this paper we address this problem by introducing a novel, end-to-end trainable approach providing concurrent object tracking and recognition. It takes as input a stream of raw sensor scene observations that is often incomplete due to occlusions, and continuously provides estimates of the uncovered, occlusion-free scene containing information about the positions of all of the objects along with their classes as illustrated in Figure~\ref{fig:intro}. 

\begin{figure*}[t]
\centering
\includegraphics[width=18cm]{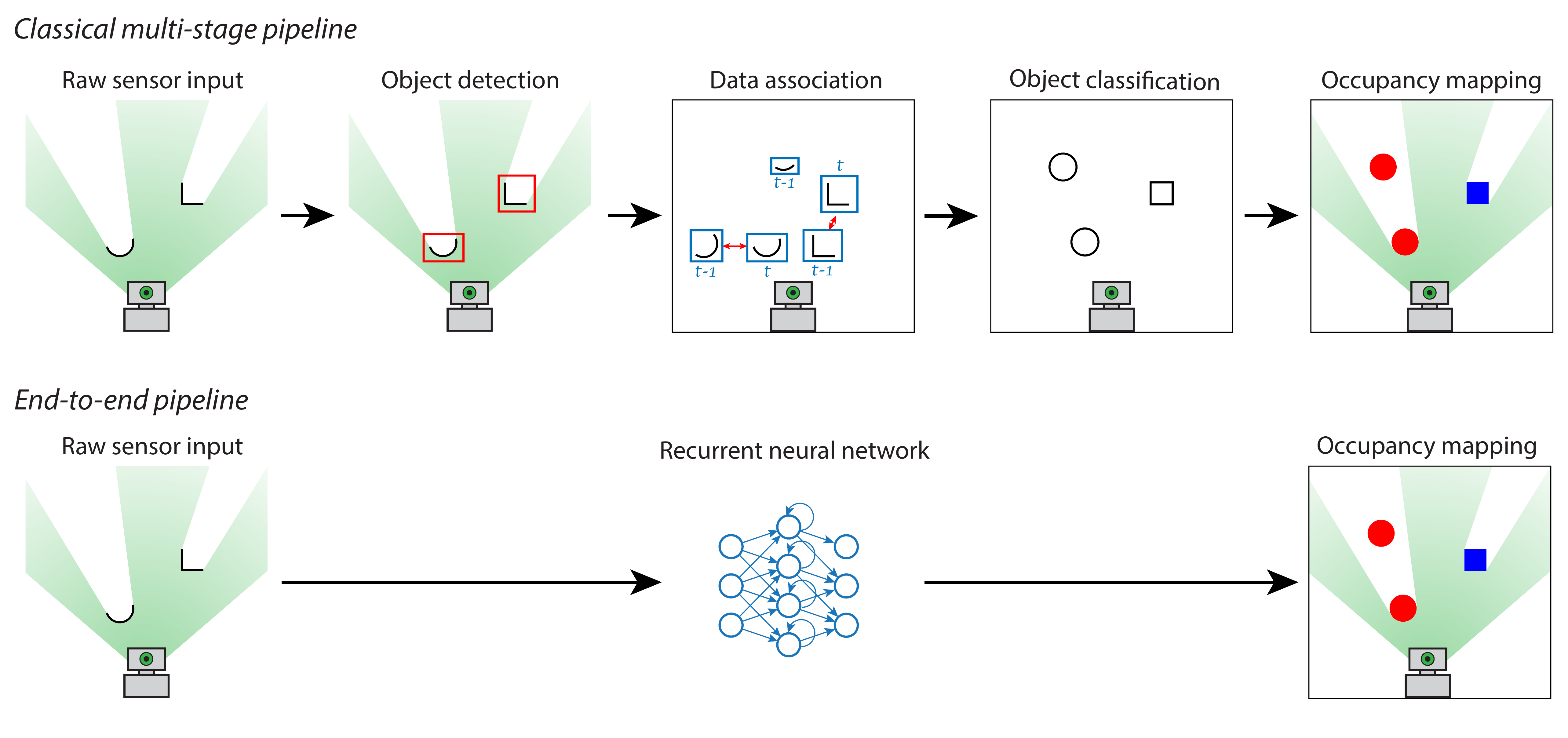}
\caption{Comparison of a classical multi-stage perception pipeline \textbf{[top]} to the proposed end-to-end framework \textbf{[bottom]}. A multi-stage pipeline requires a significant amount of design effort, and each step in the pipeline introduces its own simplifying assumptions resulting in restrictions on the general applicability of the overall system.}
\label{fig:pipeline}
\end{figure*}

In doing so, we are inspired by the recently proposed framework of \emph{Deep Tracking}~\cite{OndruskaAAAI2016} which leverages neural networks for end-to-end tracking. However, this framework has as yet only been deployed on predicting occupancy of comparatively benign, simulated data and using simple networks. We improve on this work in two ways. First, we motivate significant changes to the original \textit{Deep Tracking} architecture~\cite{OndruskaAAAI2016} and demonstrate that these lead to substantial performance gains on complex, \emph{real-world} scenarios. In particular, we propose to use multi-scale convolution to address the need to simultaneously track objects of different sizes, dynamic memory to effectively remember information for long periods of time, and static memory to learn place-specific information. Secondly, to effectively learn object semantics, we extend the framework by leveraging \textit{inductive transfer}~\cite{pan2010survey} of knowledge gathered during the tracking task to efficiently -- in terms of annotated data required -- train a classifier of the tracked objects.

We demonstrate the system on data collected from a busy road intersection, and show that it provides significantly more accurate scene prediction compared to alternative approaches
The network not only tracks and classifies different objects through complete occlusion, but also predicts their movements in a short time horizon.

The main contributions of this paper are:
\begin{itemize}
    \item A framework to allow end-to-end simultaneous tracking and semantic classification based on the method of Deep Tracking and principle of transfer learning.
    \item A tailored recurrent neural network architecture to enable tracking and semantic classification in complex, dynamic, \emph{real-world} scenarios.
\end{itemize}

The rest of the paper is structured as follows. After reviewing related works in Section~\ref{sec:litreview}, we present the problem definition in Section~\ref{sec:definition}. Section~\ref{sec:deeptracking} provides an overview of the Deep Tracking framework and Section~\ref{sec:semantic} extends the framework beyond tracking to additionally produce semantic labels of the output. Section~\ref{sec:NovelDT} proposes a new architecture to allow effective tracking in complex, real-world scenarios. Finally, in Section~\ref{sec:results}, we present an empirical evaluation of our contributions. We conclude in Section~\ref{sec:conclusion} and discuss our findings.

\section{Related Works}
\label{sec:litreview}

In this work we address the problem of effectively tracking the state of the environment around the robot. Classical approaches to this problem such as~\cite{ mertz2013moving, Wang01062015,yilmaz2006object} typically involve a multi-stage tracking pipeline as illustrated in the top row of Figure~\ref{fig:pipeline}. This pipeline features a sequence of explicit and largely hand-engineered steps consisting of object detection considering a stream of sensor input, semantic classification, data association, state estimation (including motion modelling) and, finally, occupancy grid generation. Instead, here we build on the recently proposed approach of \textit{Deep Tracking}~\cite{OndruskaAAAI2016} featuring a recurrent neural network which directly maps from raw laser data to a semantically annotated and unoccluded occupancy grid. This is illustrated in the bottom row of Figure~\ref{fig:pipeline}.

Deep learning approaches have been successful in a number of domains (see, for example,~\cite{krizhevsky2012imagenet,dahl2012context,wang2012end}) where they have benefited from large amounts of data in order to learn appropriate internal representations leading to significant performance gains above and beyond that achievable by classical methods. In our case the neural network is trained end-to-end to predict space occupancy and semantic labels directly from the raw laser data. While doing so it learns to perform an \textit{implicit tracking} where the optimal internal representations about the hypotheses of moving objects and respective update procedures of classical tracking are inferred directly from the data.

To successfully apply deep learning, an appropriate neural network architecture for the task must be chosen. Abundant literature exists on the topic of finding optimal architectures for different tasks such as convolutional networks for image processing~\cite{simard2003best} and recurrent neural nets such as \textit{long short term memory}~\cite{hochreiter1997long} or \textit{gated recurrent units}~\cite{cho2014learning} for processing sequences. We propose a novel neural network architecture specifically tailored to real-world object tracking. 
The network shares similarity with architectures for semantic labelling of natural images~\cite{yu2015multi} in terms of the ability to produce output of the same resolution. In addition we provide effective mechanisms to track objects of different sizes over time, learn place-specific information and recurrent mechanisms to remember information for long periods of time in order to track objects effectively even through long occlusions.

A common drawback of deep learning approaches is the need for large amounts of supervised data for training. We show that our network can be, in fact, trained very efficiently. The network first learns to track by just observing a stream of raw unlabelled sensor data and by trying to predict the next input. In turn we exploit the fact that the learned representation captures latent higher-order information in the data such that we can easily infer semantic labels for the tracked objects, using only a small amount of labelled data. This is a form of \emph{inductive transfer} of knowledge between machine learning tasks~\cite{pan2010survey}. In the context of neural networks it was successfully applied to a range of tasks, in the areas of multi-task learning~\cite{mitchell1993explanation,caruana1995learning} and in the form of unsupervised pre-training and supervised fine-tuning~\cite{pennington2014glove,le2013building}.

\section{Problem Formulation}
\label{sec:definition}

The input to our problem is a sequence of \emph{partially observed} states of the world, computable directly from raw sensor measurements. We represent this state as a discretised 2D grid of size $M\times M$, parallel to the ground, built locally around the sensor. The partially observed state of the world is represented by two $M\times M$ binary matrices, collectively referred to as $x_t~\in~\{0,1\}^{2 \times M \times M}$. The first matrix encodes whether a cell is directly observable (value of $1$, $0$ otherwise) by the sensor at time $t$, while the second matrix encodes whether a cell is observed to be free (a value of $0$), or occupied (a value of $1$). 
The output we wish to obtain consists of two parts. The first part is an \emph{occlusion-free} state of the world $y_t~\in~\{0,1\}^{M \times M}$, represented by an occupancy matrix similar to the occupancy matrix in $x_t$. The second part is a semantic map $c_t~\in~\{1,...,K\}^{M \times M}$, revealing, for each cell, which of $K$ types of objects (such as \emph{pedestrian}, \emph{bicyclist}, \emph{car} etc.) is currently occupying it\footnote{Whereas $c_t$ is modelled for all the cells in practice, it is ignored for cells that are not occupied i.e. $y^i_t = 0$.}.

The problem therefore resolves to solving for $P(y_t,c_t|x_{1:t})$, the probability of the complete state of the world and its semantics at time $t$, given the observed input at all previous time steps $x_{1:t}$. This formulation can also be used to predict a future state $P(y_{t+n}, c_{t+n}|x_{1:t})$ by simply providing an empty input for $x_{t+1:t+n}= \diameter$.
% An typical input and output example from a real-world scenario is shown in Figure \ref{fig:intro}.

In the next section we first outline a solution to the partial problem of estimating $P(y_t | x_{1:t})$ as suggested by the recently proposed Deep Tracking framework~\cite{OndruskaAAAI2016} but modified for operation in complex real-world scenarios. Then in Section~\ref{sec:semantic}, we extend our solution to estimate $c_t$ via the application of the principle of inductive transfer.

\section{Deep Tracking for Robotics}
\label{sec:deeptracking}

In this section, we focus on solving the first part of the problem formulated in Section~\ref{sec:definition}, namely uncovering the full, unoccluded state $y_t$ of the environment from the sequence of partially observed states $x_{1:t}$.
We use the recently proposed Deep Tracking framework~\cite{OndruskaAAAI2016} to solve this problem. However, so far, this framework has been demonstrated only on simulated scenarios composed of simple geometric objects using a simple network architecture. A number of improvements to this architecture are needed to scale up its capacity to deal with complex, dynamic, real-world data encountered in robotics applications. We first briefly review details of the framework relevant to our application, then present our proposed improved architecture.

\subsection{A Brief Review of Deep Tracking}
\label{sec:review}
Deep Tracking~\cite{OndruskaAAAI2016} is a method to model $P(y_t|x_{1:t})$ using a recurrent neural network~\cite{medsker2001recurrent}. Motivated by Hidden Markov Models~\cite{rabiner1989tutorial}, at a time $t$ a latent state $h_t$ is assumed to capture the complete information necessary for predicting $y_t$ (e.g.\ scene appearance and dynamics, locations of all objects, their shapes, velocities etc.), thus we have
\begin{equation}
P(\state_{t}|\observation_{1:t}) = P(\state_t | \hidden_t ).
\label{eq:prediction}
\end{equation}
Evolution of this latent state, which includes propagating model dynamics and integrating new sensor measurements, is modelled by an update operation
\begin{equation}
\hidden_t = f(\hidden_{t-1},\observation_t).
\label{eq:update}
\end{equation}
The key element is that both the latent state update $f(\hidden_{t-1},\observation_t)$ and its decoding to the output $P(\state_t | \hidden_t)$ are modelled as parts of a single neural network and trained jointly. Equation~\ref{eq:update} is modelled by the forward-propagation of information through hidden layers of the network, and Equation~\ref{eq:prediction} is modelled by the decoding (output) layer (cf.\ Figure~\ref{fig:network}).
Equations \ref{eq:prediction} and \ref{eq:update} can then be performed repeatedly as a form of recurrent neural network to continuously update the hidden state $h_t$ serving as a network memory and predict $y_t$, making it suitable for online stream filtering of sensor input.

\begin{figure*}[t]
\centering
\includegraphics[width=180mm]{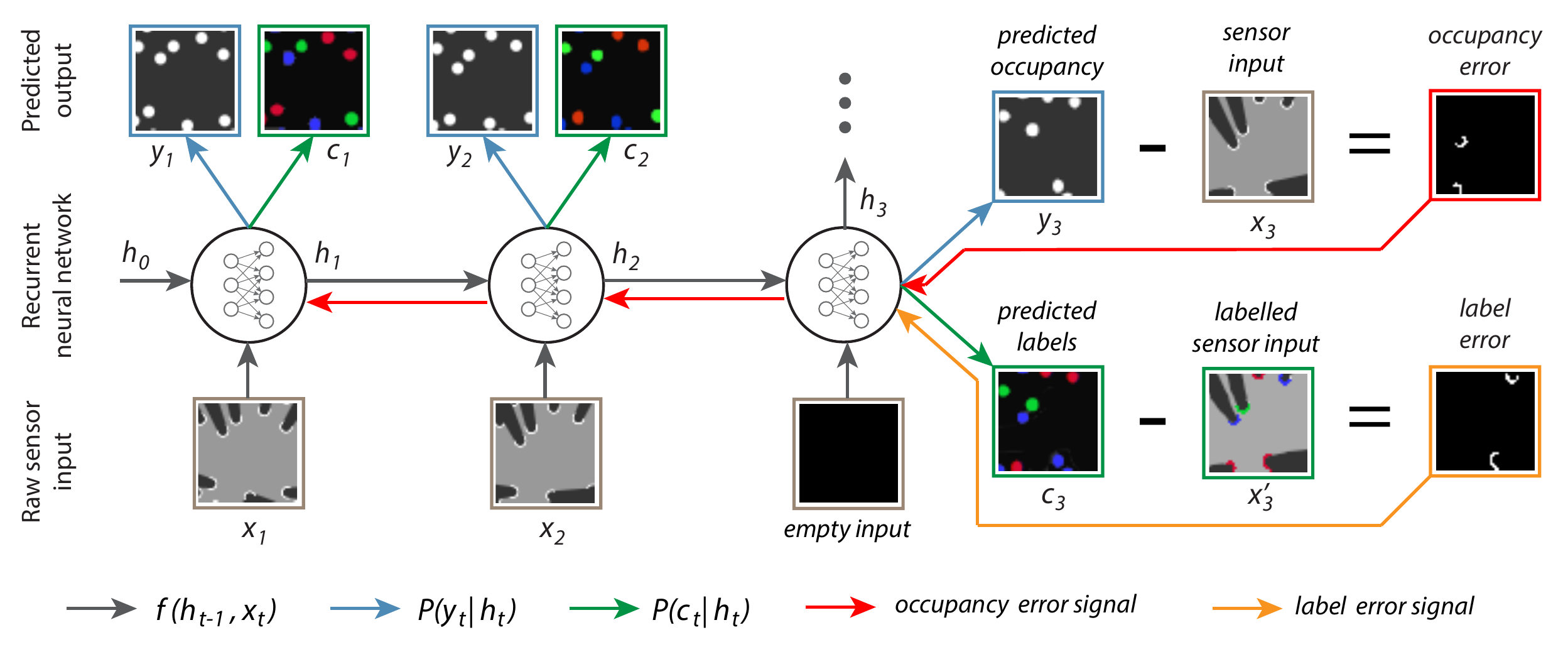}
\caption{Training of the recurrent neural network to produce both space occupancy $y_t$ and semantic labels $c_t$. The network is trained to predict output consistent with future inputs. This allows training without the need of ground-truth information of the full, unoccluded scene. First the network learns how to track by predicting correct occupancy using large amounts of unlabelled data, then a small set of labelled data is used to induce semantic classification.}
\label{fig:learning}
\end{figure*}

% \subsection{Unsupervised Training}
% \label{sub:unsupervised}

When the ground-truth output $y_t$ is not easily available, as in our case, the network can be trained in an unsupervised fashion. Here, instead of optimising directly $P(y_{t}|x_{1:t})$, the network is trained to predict $P(y'_{t+n}|x_{1:t})$ where $y'_{t+n}$ is the part of $y_{t+n}$ that is directly observed in $x_{t+n}$. This is done by predicting $P(y_{t+n}|x_{1:t})$ and back-propagating~\cite{rumelhart1988learning} the error only on the observed part of the scene (cf.\ Figure~\ref{fig:learning}). In other words, we train the network to correctly predict the subset of the ground-truth occupancy present in the \emph{future input}. As demonstrated in~\cite{OndruskaAAAI2016} and also shown in Figure~\ref{fig:results}, an important consequence of this training strategy is that, at deployment, the trained network starts to correctly imagine objects and their movement in the occluded regions. This is because the situation with occluded input at deployment is similar to that at training when no input was provided at all for the future time $t+n$, and the network was trained to predict the observable regions.

\section{Semantic Classification\\Through Inductive Transfer}
\label{sec:semantic}

In this section, we extend our solution to the partial problem $P(y_t | x_{1:t})$ presented in Section~\ref{sec:deeptracking}, to the full problem of simultaneously estimating both occlusion-free occupancy and scene semantics $y_t, c_t$. We show that this can be achieved relatively easily by exploiting the knowledge the network has already learned to predict $y_t$, through the principle of inductive transfer~\cite{pan2010survey}. The significance of this is that \emph{only a small amount} of labelled training data is needed to allow the same network to master this additional task.

The clue resides in the hidden representation $h_t$ learned in the unsupervised training for tracking, which can be viewed as a universal descriptor of the state of the world. It captures not only the positions of individual objects, but also their motion patterns, shapes and other properties necessary for the successful prediction of scene dynamics. Because the network was trained to perform well in this task a reasonable assumption to make is that any information necessary for the prediction of the position of the objects in the near future must be already contained in this hidden representation. Object semantic class falls in this category as different objects differ mainly in their shape and motion patterns. Similar to predicting $y_t$ from $h_t$ in Equation \ref{eq:prediction}, extracting $c_t$ can be achieved simply by building a classifier to predict $P(c_t|h_t)$.

\subsection{Training}
Training the classifier to extract semantic information from $h_t$ is not straightforward as a supervisory signal would need to be provided for all the pixels whether they contain an actual object or not and for both visible and occluded areas. Such a supervised dataset would be very difficult to produce if only occluded raw laser scans are available.

Instead we label only the visible cells of the available raw input data which contain an actual obstacle. Then we predict labels for all the pixels but back-propagate the error only on those with a label. This principle makes intuitive sense, however could result in the classifier to rely too much on the part of memory $h_t$ affected by presence of visible input $x_t$ and performing well only for directly visible parts of the scene but poorly for the occluded objects where the prediction is driven purely by the previously remembered information in $h_t$. To address this issue a more elaborate training procedure is necessary.

To ensure good performance of the network on classifying \emph{occluded} objects it must be trained in such settings. This can be achieved using the same principle used to train the network to predict $y_t$ from $h_t$. We train the network to predict the future semantic label $c_{t+n}$ while providing only input $x_{1:t}$ which forces the network to use information stored in memory $h_t$ and then back-propagate the error compared to the true label of $x_{t+n}$.
The entire process is illustrated in Figure~\ref{fig:learning}.

\section{The Network Architecture}
\label{sec:NovelDT}

\begin{figure}[b]
\centering
\includegraphics[width=80mm]{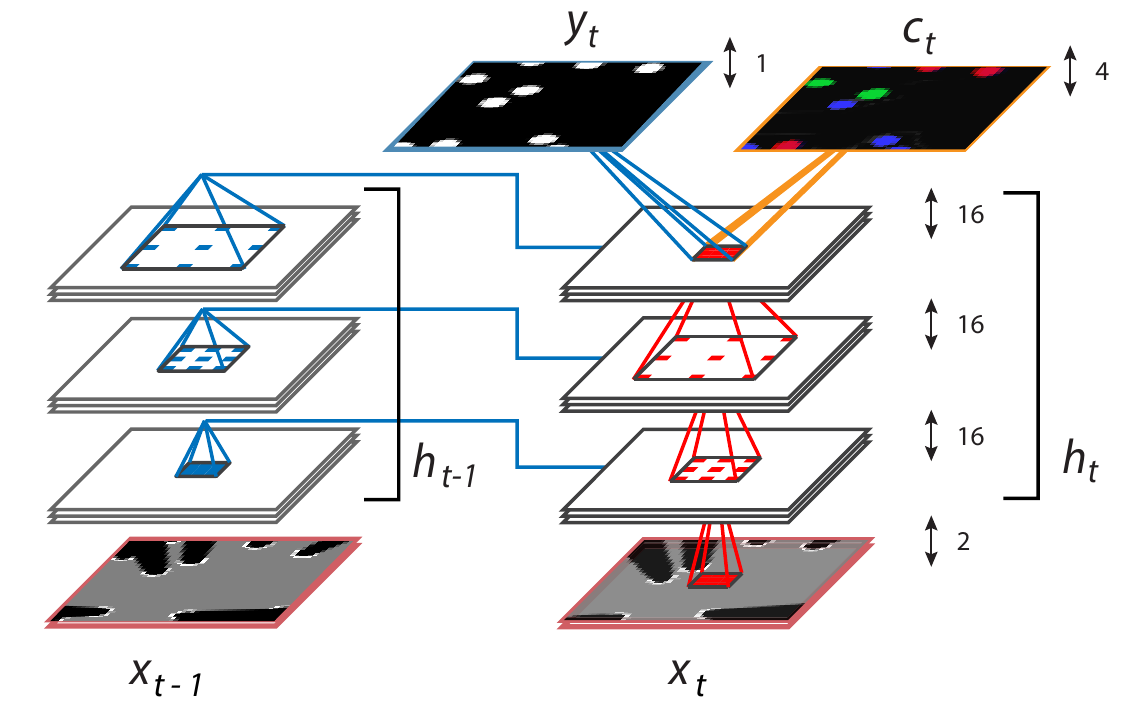}
\caption{The proposed architecture for tracking and semantic classification. It features dilated convolution, enhanced static and dynamic memory capabilities whereas producing information of both cell occupancy and it's semantic class.}
\label{fig:network}
\end{figure}

The simple recurrent neural network proposed in~\cite{OndruskaAAAI2016} was demonstrated to be sufficient for the simulated dynamic scenario evaluated in that work. However, an effective deployment in real-world robotics applications poses a set of challenges for which a more appropriate architecture must be chosen. In particular it requires the ability to simultaneously track objects of different sizes such as cars and pedestrians, to effectively remember information for long periods of time to deal with occlusions, learn and exploit place-dependent information such as the presence of static obstacles and lastly produce output for both space occupancy and it's label. We therefore in this section present a new network architecture designed to address the above issues.

An overview of our proposed network is depicted in Figure~\ref{fig:network}. 
The input $x_t$ at time $t$ is processed by a multi-layer network. At each layer the output of the previous layer is combined with it's own activations at time $t-1$ implementing the recurrence. This allows the network to extract and remember the information from the past and use it for prediction at time $t$. The output of the final layer is then converted into the resulting output $y_t$ through simple convolutional decoder.

Unlike classical convolutional networks such as~\cite{Simonyan14c} this network \textit{does not} feature max pooling and maintains the same resolution in each layer.
In addition it features four key elements critical for successful tracking and classification in realistic scenarios: \emph{multi-scale convolution}, \emph{dynamic memory}, \emph{static memory}, and \emph{pair of decoders} which we describe below.

\begin{figure}[t]
\centering
\includegraphics[width=80mm]{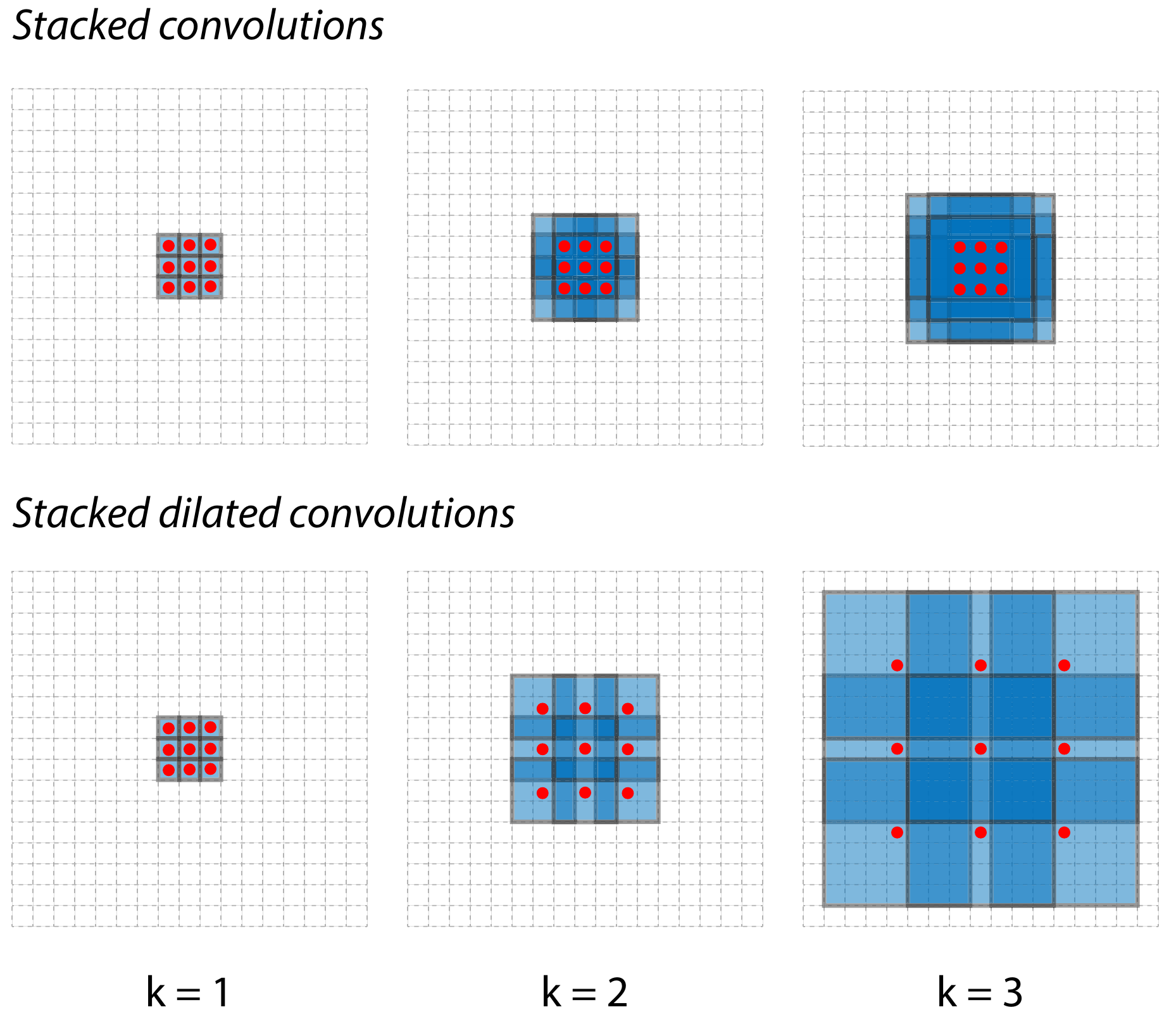}
\caption{Multi-scale context aggregation preserving the image resolution by stacking \textit{dilated convolutions}~\cite{yu2015multi}. At layer $k$ the red pixels are convolved with a skip of $2^{k-1}-1$ pixels. This results in exponential growth of the blue \textit{receptive field size} \textbf{[bottom]} as opposed to stacking classical convolutions resulting only in linear growth \textbf{[top]}.}
\label{fig:dilating}
\vspace{-3mm}
\end{figure}

\subsubsection{Multi-Scale Convolution}

For the network to correctly predict the occupancy and label at location $i$, such as affected by the presence of a moving object, this object must fall in the \textit{receptive field} of the neuron in the final layer. The \textit{receptive field} is the part of the input affecting the value of the neuron. In the case of classical convolution the \textit{receptive field} is the $K \times K$ neighbourhood where $K$ is the size of the convolution kernel. The size of the receptive field however limits the size of effectively tractable objects in the input which can be of vastly different sizes in realistic settings. One way to increase the receptive field is to increase the kernel size or stack multiple convolutions on top of one other. This however creates a computational challenge as the number of parameters and computational complexity grows quadratically with $K$ in the first case and linearly in the second case.

Instead we use a stack of dilated convolutions~\cite{yu2015multi} where the receptive field grows exponentially with the number of layers. The basic idea is to perform the classical $3 \times 3$ convolution but skipping $2^{k-1}-1$ pixels in between convolved pixels at layer $k$, as illustrated in Figure~\ref{fig:dilating}. This gives a $(2^{K+1}-1) \times (2^{K+1}-1)$ receptive field at final layer $K$. This dilated convolution is then used as an elementary computation step to implement the \emph{dynamic memory} described below.

\subsubsection{Dynamic Memory}

To be able to track a moving object through extended periods of occlusion, the network must remember the location of the object and other properties such as its shape and velocity. Findings from studies on recurrent neural networks stress out the importance of specially dedicated units such as \textit{long short term memory}~\cite{hochreiter1997long} to support information-caching, as otherwise training suffers from the vanishing gradient problem~\cite{pascanu2012difficulty}. Inspired by~\cite{xingjian2015convolutional}, we implement a convolutional variant of \textit{gated recurrent units}~\cite{cho2014learning} as the processing step at each layer. The output of each unit is given by the weighted combination of its previous output at time $t-1$ and a candidate memory $\bar{h}_t$ computed from the output of the layer bellow with the forgetting of information controlled by the reset gate $r_t$:
\begin{eqnarray}
\label{eq:ConvGRUz}
f_t &=& \sigma(W_{xz} \ast x_t + W_{hz} \ast h_{t-1} + b_z)\mbox{ ,}\\
\label{eq:ConvGRUr}
r_t &=& \sigma(W_{xr} \ast x_t + W_{hr} \ast h_{t-1} + b_r)\mbox{ ,}\\
\label{eq:ConvGRUh}
\bar{h}_t &=& \tanh(W_{xh} \ast x_t + r_t \circ W_{hh} \ast h_{t-1} + b_h )\mbox{ ,}\\
h_t &=& f_t \circ h_{t-1} + (1 - f_t) \circ \bar{h}_{t}\mbox{ .}
\end{eqnarray}
Here $*$ denotes dilated convolution described earlier and $\circ$ denotes element-wise multiplication.

\subsubsection{Static Memory}
 We allow each cell to learn a unique and universally accessible piece of information different from all other cells. This is achieved by biases $b_z$, $b_r$, and $b_h$ in Equations~\ref{eq:ConvGRUz}-\ref{eq:ConvGRUh} which are not a per-layer constant as in the case of classical convolution, but are learned individually for each neuron during the training. As shown in Section~\ref{sec:results} this allows the network to learn place-specific information such as the static occupancy of the cell or the usual motion patterns and classes in a particular area, which can then be used to aid the network prediction.

\subsubsection{Decoders}
Finally, we employ pair of simple convolutional decoders to decode output of the final layer to the cell occupancy $y_t$ and class label $c_t$. The difference is that for the class label we employ softmax (multinomial logistic regression) output for the $K$ classes instead of a sigmoid function.

\section{Results}
\label{sec:results}

\begin{figure}[b]
\centering
\includegraphics[width=85mm]{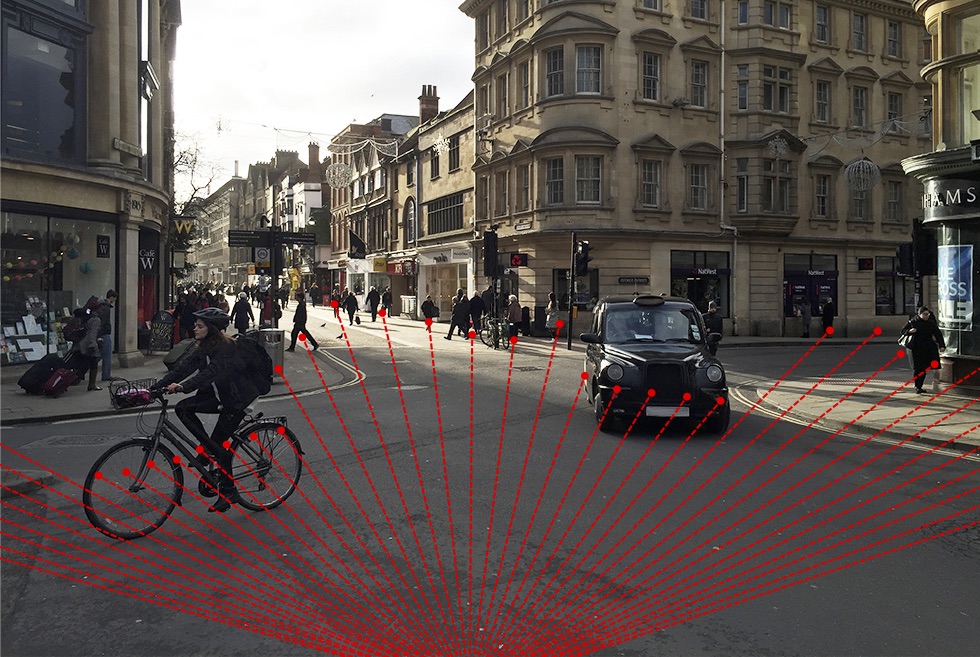}
\caption{Location of the experiment from the robot's point of view with a superimposed illustration of laser measurements. The area is occupied by a variety of different dynamic objects such as pedestrians, cyclists and cars.}
\label{fig:location}
\end{figure}

In this section we demonstrate the efficacy of the proposed system in both tasks of tracking and semantic classification in a complex, real-world scenario. We show that the trained network is able to track and classify a variety of different objects even through complete occlusion, and is able to predict the evolution of the scene in the near future. In particular, we are interested in the relative performance of the proposed network architecture in correctly predicting both  object positions and class labels. 
We show that the achieved tracking accuracy is superior to the original architecture presented in~\cite{OndruskaAAAI2016}, as well as to that of an alternative state-of-the-art model-free tracking solution targeted at the same problem~\cite{Wang01062015}.

\begin{figure}[t]
\centering
\includegraphics[width=75mm]{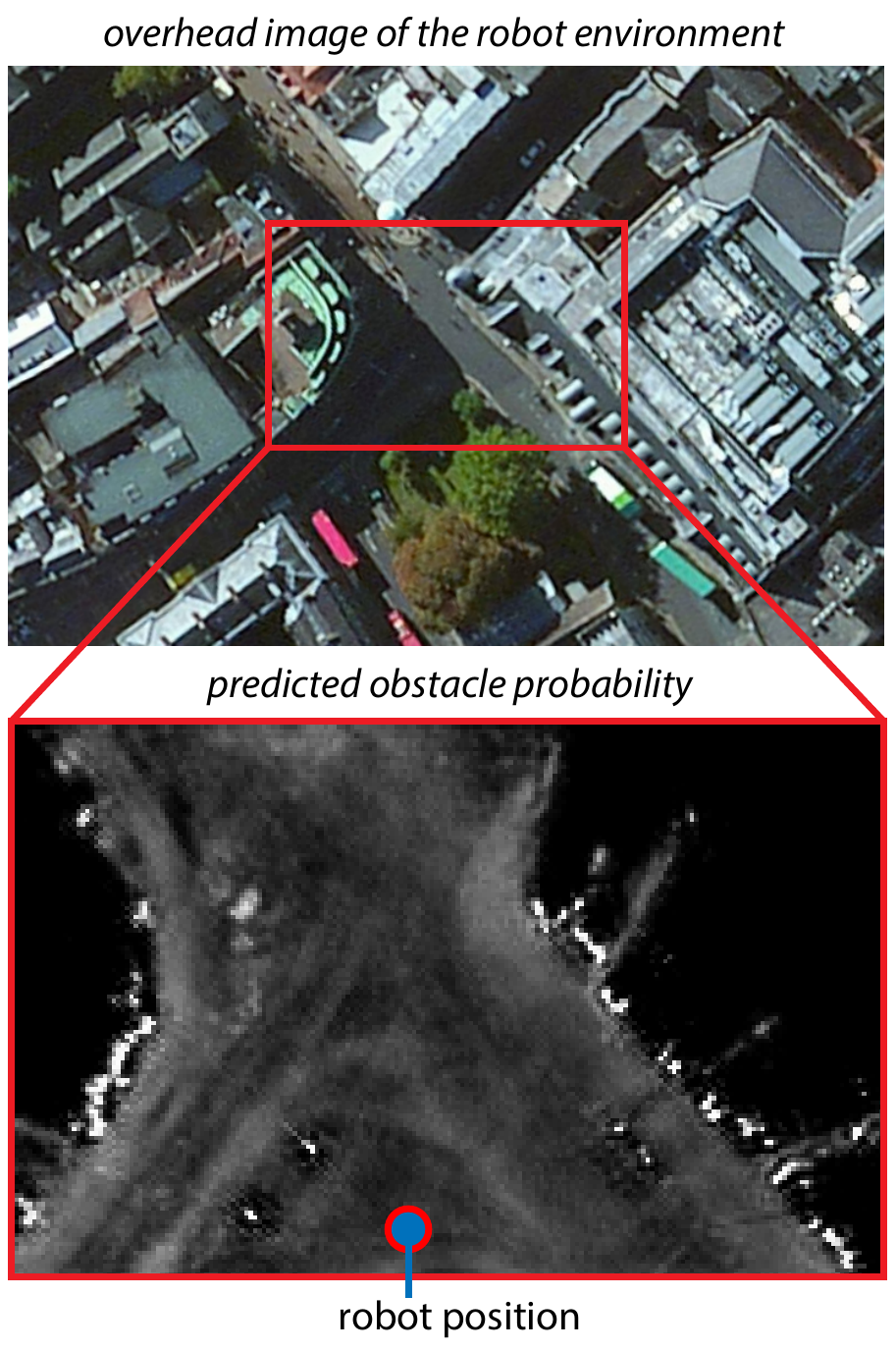}
\caption{Trained network output when provided no input \textbf{[bottom]} and corresponding aerial view of the robot environment \textbf{[top]}. The ability of the network to learn per-pixel information allows adaptation to the training environment. This allows the network to confidently predict the position of static obstacles such as buildings, as well as the probability of any given cell being occupied even without any sensor input. Pavements show higher probabilities than the centre of the roads. For clarity of visualisation, we show here the log of the probabilities of occupation.}
\label{fig:static}
\end{figure}

\subsection{Dataset}
We collected a 75 minute long log from a stationary robotic platform equipped with a \textit{Hokuyo UTM-30LX} 2D laser scanner, positioned in the middle of a busy urban intersection, as depicted in Figure~\ref{fig:location}. The area features dense traffic composed of buses, cars, cyclists and pedestrians, which results in extensive amounts of occlusion. Consequently, at no point in time the complete scene was fully observable. We subsampled the dataset at 8Hz and split it into a 65 minute unsupervised set to train the network, and a 10 minute long test set to measure the occupancy prediction performance. In addition, we hand-labelled 800 scans from the training set into 4 classes for the purpose of network semantic training, and 200 scans from the test set for the evaluation of its semantic classification accuracy. The classes considered are: \emph{pedestrian}, \emph{car/bus}, \emph{cyclist} and \emph{background} (static obstacle). 

The input to the network $x_t$ (the partially observed occupancy grid) is computed from raw 2D laser scans by ray-tracing. Cells where a laser measurement ends are marked as occupied, all cells from the sensor origin up to the end of the ray are marked as free, and cells beyond the ray are marked to be unobserved. We build a $100~\times~100$ grid around the sensor, and each cell covers a $20~\times~20\mbox{ cm}^2$ area, the input grid thus spanning a total area of $20 \times 20 \mbox{ m}^2$.

\begin{figure*}[t]
\centering
\includegraphics[width=180mm]{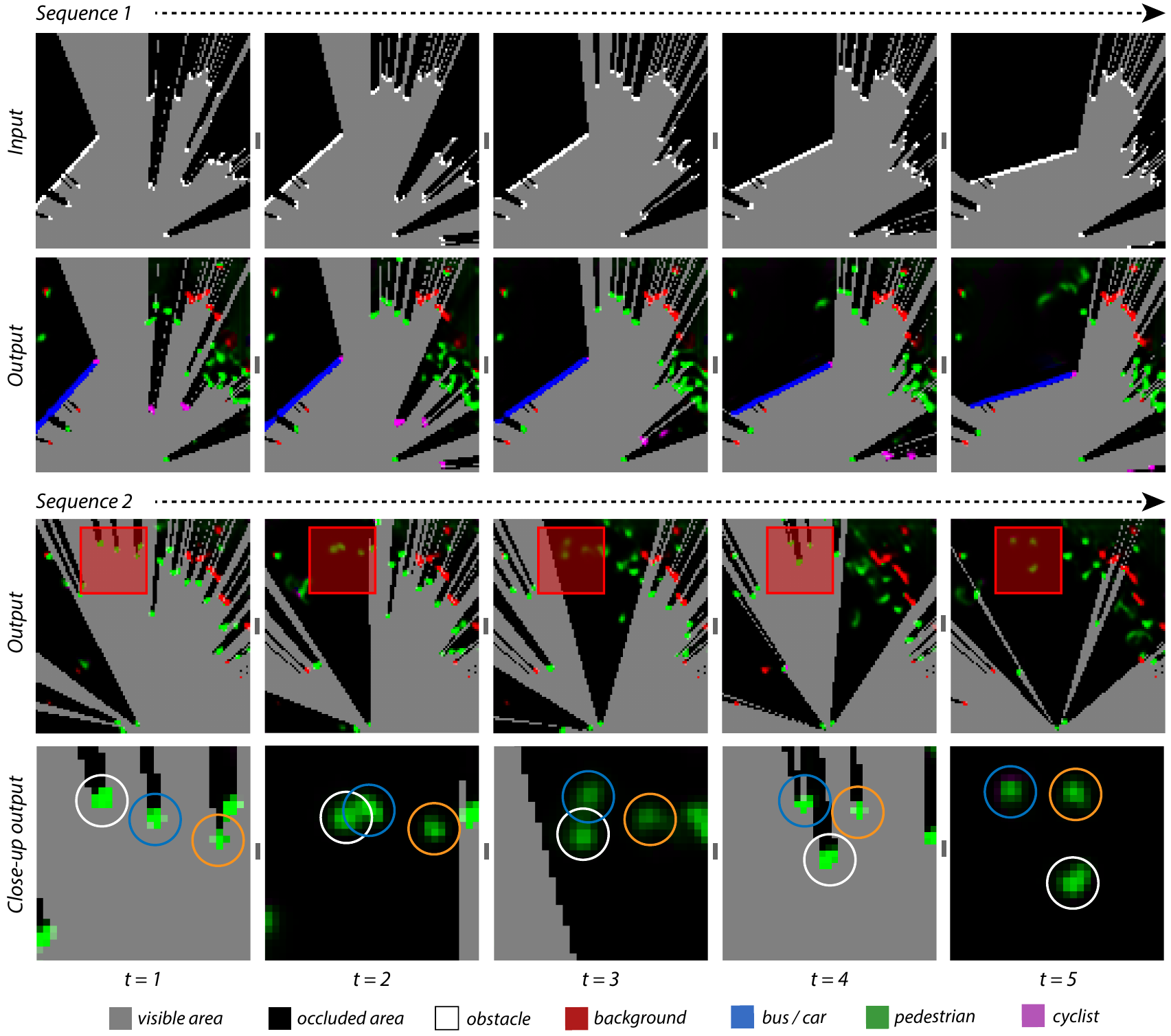}
\caption{Example of produced outputs of the system. As displayed in the highlighted close-up of the output of second sequence the network is able to propagate assumed motion of the objects (highlighted by circles) even when in complete occlusion. }
\label{fig:results}
\end{figure*}

\subsection{Network training}
We tested the proposed network architecture described in Section~\ref{sec:NovelDT} configured with three hidden layers of 16 channels per layer. This network has a total number of 1,010,193 parameters and was trained for 72 hours until convergence on a single Nvidia Titan GPU with 6 GB of memory, using the unsupervised training procedure described in Section~\ref{sec:review}. The training sequence was split into mini-batches of length 40 (5 seconds of stream). For every mini-batch, the network is shown 10 frames and trained to predict the next 10 frames, leading to two such sequences per mini-batch. This was chosen to cover the usual length of the occlusions in the scene and we expect it would need to be increased for longer-lasting occlusions. 

Next, the classifier of $P(c_t|h_t)$ was trained in just 2 hours on the semantic classification task using the 2 minute long labelled dataset. Because the dataset is skewed and contains many more objects of a particular type, e.g.\ pedestrians, we used a weighting scheme assigning class weights equal to the inverse of class frequency.

\begin{figure}[b]
\centering
\includegraphics[width=80mm]{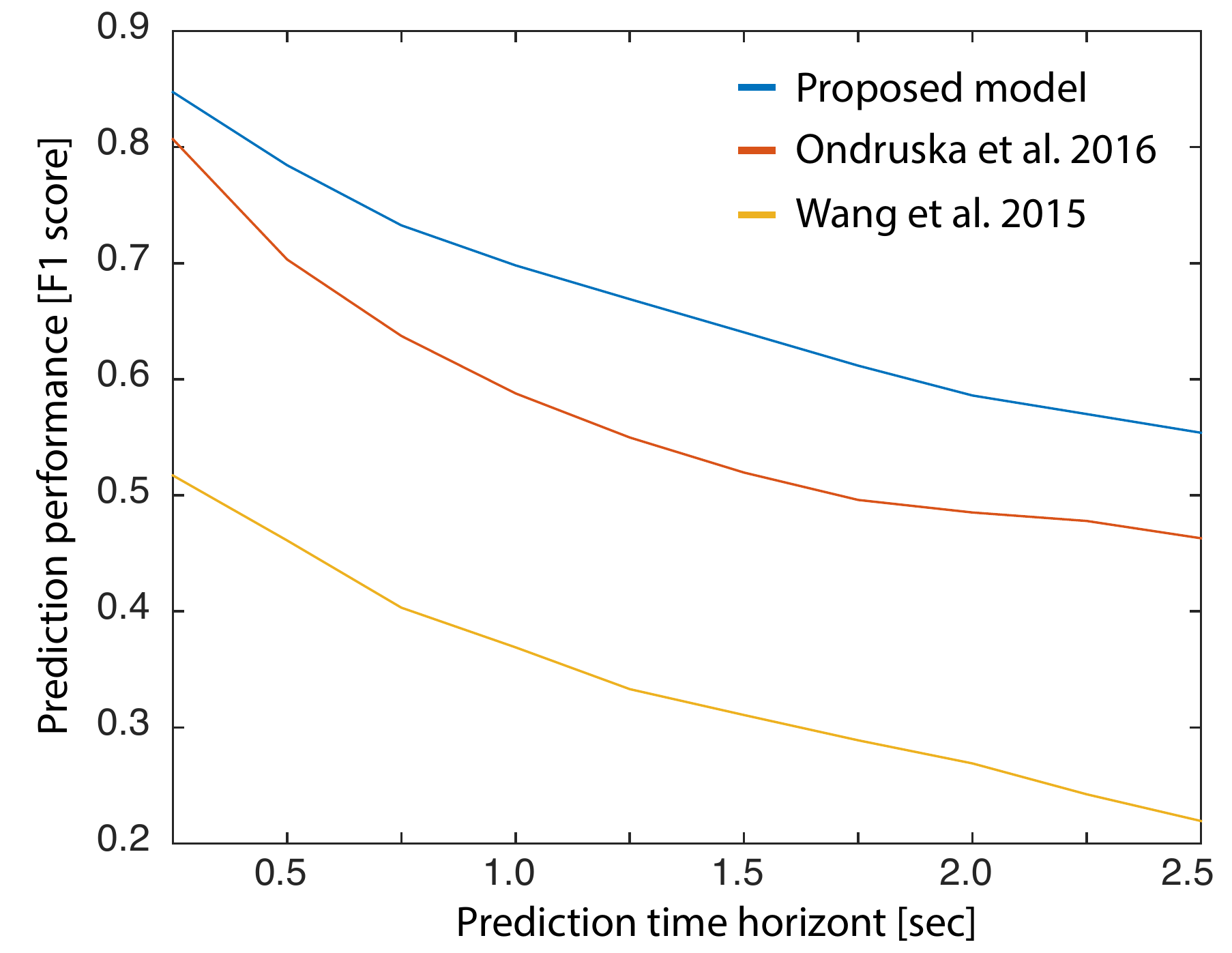}
\caption{Network ability to correctly predict the future occupancy of the scene in the time horizon of 2.5 seconds, measured by consistency with the future input. As the time horizon increases the quality of prediction degrades. The proposed neural network architecture performs better than the original architecture presented in~\cite{OndruskaAAAI2016} as well as a state-of-the-art multi-stage pipeline approach~\cite{Wang01062015}.}
\label{fig:curve}
\end{figure}

\begin{figure}[t]
\centering
\includegraphics[width=85mm]{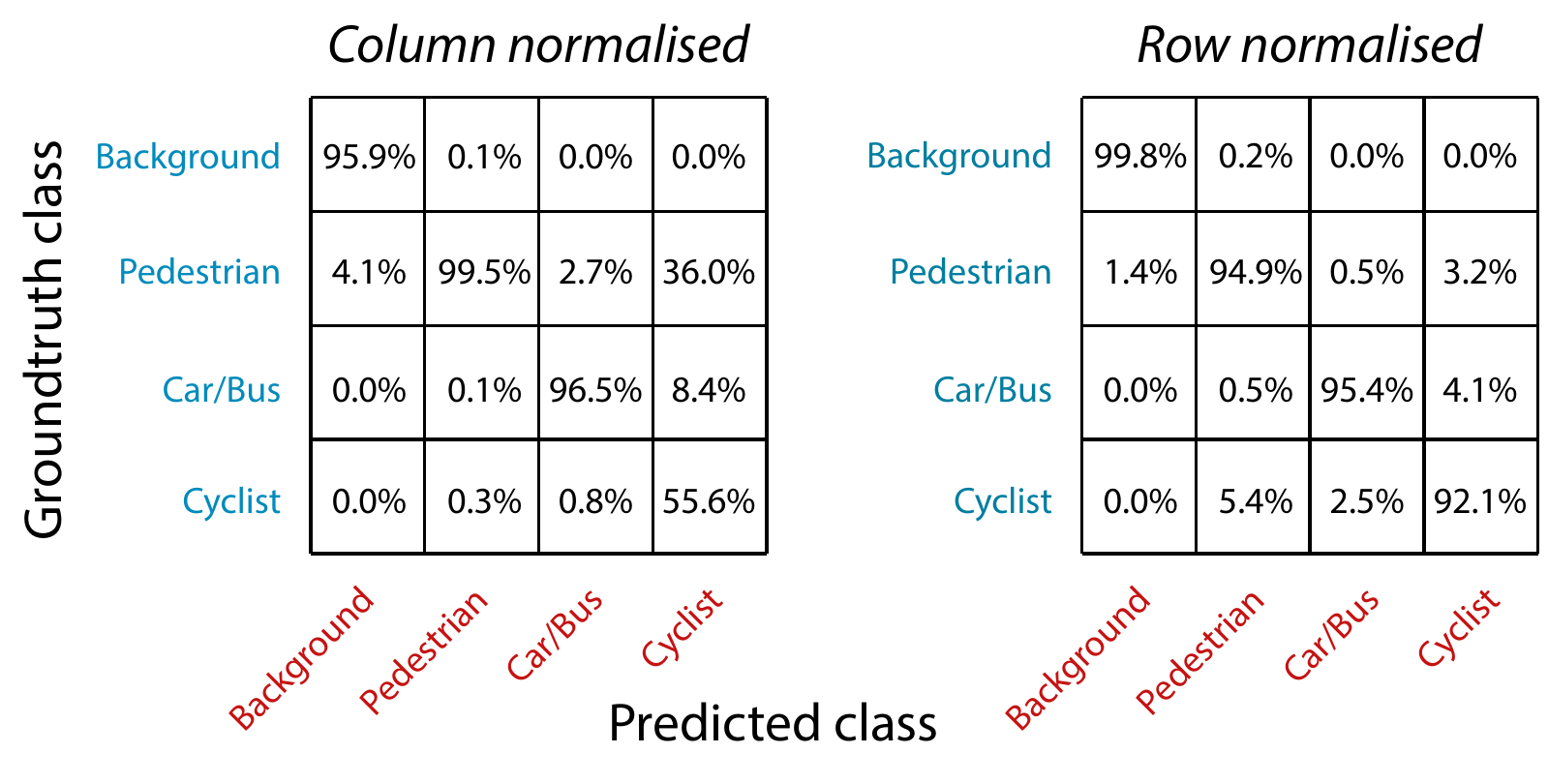}
\caption{The confusion matrix of the semantic classification performance. }
\label{fig:matrix}
\end{figure}

\subsection{Benchmarking Against an Existing Approach}
To compare the performance of the proposed end-to-end system to more traditional multi-stage pipelines, we evaluate the ability to predict future movement of dynamic objects of the proposed framework against a recently proposed state-of-the-art approach~\cite{Wang01062015} based on model-free tracking of dynamic objects using a Kalman filter. This method accepts raw laser scans, performs data clustering and association, as well as velocity estimation of moving objects. This information is then used to predict the positions of individual points in the future. We tuned the parameters of this method on the training set, then collected results from the test set, and converted the output into occupancy grids for comparison with the proposed method.

\subsection{Evaluation}
Two typical input sequences and their corresponding predicted network output are shown in Figure~\ref{fig:results}. The network is able to uncover the unoccluded scene including the space occupancy $y_t$ and object labels $c_t$. Moreover it is able to update the positions of dynamic objects through temporary occlusion demonstrating that it has learned to track and recognise objects in the scene. 

In what follows, we quantitatively evaluate the performance of the proposed system to justify these qualitative observations.

\subsubsection{Occupancy Accuracy}

As the ground-truth occupancy annotations $y_t$ of the full unoccluded state was not available we, instead, measured the accuracy of predicting the future occupancy $y_{t+n}$ with respect to the visible part of the scene in $x_{t+n}$. This is the same metric as the one used to train the network in Section \ref{sec:deeptracking}. For binary obstacle prediction, we show in Figure~\ref{fig:curve} the computed F1-scores when predicting 10 consecutive frames, averaged over the test set, and compare to both the original architecture described in~\cite{OndruskaAAAI2016} (that we train in a similar manner), and to the state-of-the-art multi-stage pipeline approach of~\cite{Wang01062015}.
The prediction is accurate in the near horizon and progressively decreases over time. This is expected, as the uncertainty of the state of the world increases with the prediction horizon. In both cases our results outperform the two alternative approaches, demonstrating the effectiveness of the proposed network architecture and advantages of end-to-end learning.

Another experiment is concerned with measuring the effectiveness of the desired ability to learn place-specific information. One way to evaluate this is to visualise the network prediction $y_1$ without providing it with any input as displayed in Figure~\ref{fig:static}. Even without input sensor information, the network is able to provide an estimate of the expected occupancy probabilities, which is higher at the locations of static obstacles and at crowded areas of the scene such as pavements. As no propagation of the information through the network occur this is clearly only made possible by the ability of the network to remember this information in its static memory during training.

\subsubsection{Semantic Accuracy}
To quantify the network's ability to classify scene semantics,
we compute the confusion matrix which is shown in Figure~\ref{fig:matrix}. As can be seen, the network is able to produce reliable classification for the object classes considered. The main source of error lies in distinguishing cyclists from pedestrians, as they often exhibit similar shapes in 2D laser data.

To verify the value of the proposed inductive transfer of knowledge, we compare this result to an alternative approach of classifying scene semantics, which takes the form of a one-shot classification of all directly visible obstacles from a single raw-sensor input. As a representative solution to such an approach, we trained a three-layer deep convolutional classifier predicting $c_t$ directly from $x_t$. Despite conducting rigorous parameter tuning, classifying directly the input $x_t$ yields inferior classification accuracy compared to classifying the hidden representation $h_t$, the negative log likelihood of correct labels being respectively $\sum_t -\log P(c_t|x_t) = \mathbf{101.967}$ and $\sum_t -\log P(c_t|h_t) = \mathbf{49.129}$. This demonstrates that $h_t$ offers a powerful semantic descriptor of the scene and can be used as input for accurate semantic classification.

\subsubsection{Timing}
The forward propagation of a single input through the network takes 15ms on an Nvidia Titan GPU and 83ms on a commodity laptop CPU. This is sufficient to enable real-time processing of the considered stream of laser data at 8Hz.

\section{Conclusions}
\label{sec:conclusion}

In this work, we presented a novel end-to-end trainable solution for real-time object tracking and classification in complex and partially-observable real-world environments. Leveraging the representational power of recurrent neural networks and employing efficient training procedures, the method surpasses a representative state-of-the-art model-free method, while substantially reducing the requirement for hand-engineered knowledge.

The method can be extended or applied in a number of ways. The universal schema of input and output opens up a possibility to apply the method to situations beyond those evaluated, such as accounting for robot motion which can be handled by moving the robot inside the grid, or multi-sensor or multi-robot fusion. Additionally, the knowledge of the environment captured in learning to track can be further exploited to provide different kinds of semantic information such as bounding boxes. Finally, the inherent ability to predict the future evolution of the environment around the robot can be leveraged upon in more far-sighted planning scenarios.

\bibliographystyle{plainnat}
\bibliography{main}

\end{document}